# End-to-end Multimodal Emotion and Gender Recognition with Dynamic Joint Loss Weights

Myungsu Chae¶†, Tae-Ho Kim§†, Young Hoon Shin§, June-Woo Kim§, Soo-Young Lee§

*Abstract*—Multi-task learning is a method for improving the generalizability of multiple tasks. In order to perform multiple classification tasks with one neural network model, the losses of each task should be combined. Previous studies have mostly focused on multiple prediction tasks using joint loss with static weights for training models, choosing the weights between tasks without making sufficient considerations by setting them uniformly or empirically. In this study, we propose a method to calculate joint loss using dynamic weights to improve the total performance, instead of the individual performance, of tasks. We apply this method to design an end-to-end multimodal emotion and gender recognition model using audio and video data. This approach provides proper weights for the loss of each task when the training process ends. In our experiments, emotion and gender recognition with the proposed method yielded a lower joint loss, which is computed as the negative log-likelihood, than using static weights for joint loss. Moreover, our proposed model has better generalizability than other models. To the best of our knowledge, this research is the first to demonstrate the strength of using dynamic weights for joint loss for maximizing overall performance in emotion and gender recognition tasks.

## I. Introduction

Accurate emotion and gender recognition from human conversations is one of the key challenges when developing artificial intelligence (AI) assistance systems. Because most commercial off-the-shelf AI assistance systems only focus on the users' speech and do not consider their emotions, it is difficult for them to find the users' exact intention. Thus, it is necessary for AI assistance systems to recognize humans' emotion and gender in order to communicate with them.

There are two different ways to represent emotions: discrete and continuous representations [1]. Discrete approaches consist in classifying emotions into several categories, such as the Big Six emotions [2]. On the other hand, continuous approaches present ways to represent emotions along various continuous dimensions, such as arousal and valence [3]. In this paper, the former approach is applied.

Previous works on emotion and gender recognition mainly focused on extracting meaningful features [4], [5], [6]. Genders can be recognized relatively more easily than emotions [7], [8] in general. Therefore, previous studies mostly concentrated on emotion recognition rather than gender recognition. Since 2016, the studies have proposed end-to-end approaches for an emotion recognition system that does not use any pre-trained features based on the user's speech [9] and face [10]. However, for accurate emotion recognition, it is essential to consider the various modalities of human communication, such as speech, video, text, bio-signals, etc.

Many previous researchers used joint-loss concepts, which are commonly applied to multi-task learning, for the prediction of discrete classes of emotion [12], [17]. In multi-task learning, the performance of a model is highly dependent on the joint loss weights of each task, so selecting proper weights is an important issue for multi-task learning approaches. One method to determine the joint loss weights consists in using prior knowledge of models [12]. However, these static weights can cause adverse effect because each of the loss values continuously changes while training the models. Therefore, we propose an end-to-end multimodal emotion and gender recognition model using joint loss with dynamic weights.

The rest of this paper is organized as follows. Section II presents previous works related to this research field. Section III provides an overview of the dataset used in the experiments. Section IV introduces the main methodology of the proposed approach. Section V presents an experimental analysis for comparing the performance of the proposed model and that of the baseline approach. Lastly, Section VI presents a summary and the future direction of this work.

## II. Related Work

A number of studies about emotion and gender recognition used pre-defined features, such as the mel-frequency cepstral coefficient (MFCC) from speech and geometric features from video. Feature extraction from speech and video for emotion and gender recognition is not only an active research field, but also a challenging task. Even though models made are highly dependent on the extracted features, their performance for recognition tasks is improving. In [10], the first end-to-end approach was proposed, which showed better performance than other methods by using visual information from images. After this, an end-to-end

This work was supported by the Institute for Information & Communications Technology Promotion (IITP), grant funded by the Korea government (MSIT) [2016-0-00562 (R0124-16-0002), Emotional Intelligence Technology to Infer Human Emotion and Carry on Dialogue Accordingly], and by the Industrial Strategic Technology Development Program (10076757, Free-Running Embedded Speech Recognition Technology for Natural Language Dialogue with Robots) funded by the Ministry of Trade, Industry and Energy (MOTIE, Korea).

¶ Myungsu Chae was with Korea Advanced Institute of Science and Technology (KAIST), Institute for Artificial Intelligence, Daejeon, 34141 Republic of Korea. He is now with NOTA Incorporated, Daejeon, 34141 Republic of Korea. (phone: +82-42-350-7092; fax: +82-42-350-8490; e-mail: mschae89@nota.ai).

§ Tae-Ho Kim, Young Hoon Shin, June-Woo Kim, and Soo-Young Lee are with Korea Advanced Institute of Science and Technology (KAIST), Institute for Artificial Intelligence, Daejeon, 34141 Republic of Korea.

† These authors contributed equally to this work.

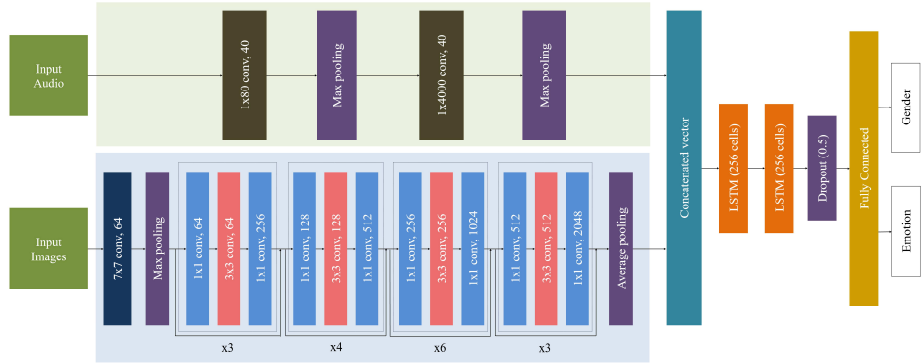

Figure 1. Network architecture of the proposed methodology

facial emotion recognition method was demonstrated in [5]. As for speech emotion recognition, the approach presented in [14] was the first end-to-end approach to use a raw voice signal as an input to neural networks for recognizing emotions. In [9], [10], and [11], the authors achieved better performance on emotion recognition than conventional approaches, which need pre-defined features.

As stated above, it is necessary to consider various modalities for accurate emotion recognition. Many previous researches applying multimodal approaches showed better performance than unimodal approaches in emotion recognition applications. In [15], the authors proposed a model using variants of a relevance vector machine for using audio and video data. Moreover, in [16], a strength-modeling framework was demonstrated, which was implemented by combining feature-level and decision-level strategies together with the conventional features for multimodal emotion recognition. A method proposed in a more recent research work described in [9], which is the baseline method used in this paper, showed better performance in multimodal emotion recognition than the methods presented in [15] and [16], which use multimodal concepts based on pre-defined features.

In multi-task learning approaches, the performance of the models highly depends on the joint loss weights. There are two common ways to select proper weights for calculating joint loss: static and dynamic approaches. Static approaches use fixed weights that are determined in advance, and thus they do not change during training. Such methods are effective for multimodal training when prior knowledge of each model is given. Other approaches use dynamic weights and do not need any prior knowledge. These dynamic weights are derived from the model during the training process. The final weights can be determined at the end of training process. Previous studies [11], [12] used static weights for calculating joint loss and selected them empirically, even though they did not have any prior knowledge about them. In this work, we propose an end-to-end multimodal emotion and gender recognition model using joint loss that combines multiple loss functions to simultaneously and adaptively learn multiple weights.

III. DATASET

In this research, the Interactive Emotional Dyadic Motion Capture database (IEMOCAP) [18] was used for our experimental analysis. The dataset contains approximately 12 hours of audio and visual data, including motion capture data of faces, video, speech, and transcription of texts. To create this dataset, a group of ten professional actors composed of five males and five females acted in two different scenarios, namely scripted play and improvised dialog. Each dialog was approximately five minutes long, but we segmented the audio and video data into utterances. Three different annotators were asked to label for nine discrete emotions (anger, excitement, happiness, sadness, neutral, frustration, fear, surprise, and disgust) and two annotators evaluated the continuous dimension of emotions, which are arousal (activation), valence, and dominance, scaled from 1 to 5. In our study, we used four discrete emotions (anger, happiness, neutral, sadness) [18], and the prediction targets were determined by taking the majority of the evaluations from the three annotators [19]. If any conflict arose when choosing the majority, the target was selected randomly.

IV. METHODOLOGY

In this section, a method for end-to-end multimodal emotion and gender recognition via deep-learning algorithms and a mechanism for determining the dynamic joint loss weights, which can minimize the negative log-likelihood of the two classification tasks, are proposed.

A. Network Architecture

Because the novelty of this research lies in that it focuses on a method for determining joint loss, the basic network architecture used is very similar to that of the baseline work [11] in order to make a fair comparison of performance. Fig. 1 shows an overview of the network architecture. It consisted of two main parts: a speech network and a visual network. In the speech network, the raw voice signal had a sampling rate of 16 kHz, and thus the input vector consisted of 96000 dimensions. We used 40 filters with a 5-ms window and a size of 80 in the first convolution layer. Then, the signal was downsampled to 8 kHz in a max-pooling layer with a pool size of 2. In the second convolution layer, we used 40 filters with a 500-ms window and a size of 4000. Afterwards, we performed max-pooling with a pool size of 10 on the channel dimension.

In the visual network, we used a deep residual network (ResNet) with 50 layers [20]. The pixel intensities of each subject's video were cropped down to 224×224, downsampled to 96×96, and used as the input of the visual network. The first part of this network consisted of a 7×7 convolution layer with 64 feature maps and a max-pooling layer with a size of 3. The second part was composed of four bottleneck architectures, which had three convolution layers of sizes 1, 3, and 1 for each residual function. At the end of the last bottleneck architecture, we performed average pooling.

TABLE 1. COMPARISON OF THE EXPERIMENTAL RESULTS FOR EACH APPROACH. THE TASK WEIGHTS OF THE DYNAMIC MODELS ARE THE FINAL VALUES AFTER TRAINING.

| Model | Modality Used | | Task Weights | | Negative Log-likelihood (validation dataset) |
|---|---|---|---|---|---|
| | Audio | Video | Emotion | Gender | |
| UM-dynamic-A | ✓ | - | 0.3232 | 0.6768 | 158.38 |
| UM-dynamic-V | - | ✓ | 0.3623 | 0.6377 | 136.08 |
| MM-static (baseline [11]) | ✓ | ✓ | 0.5 | 0.5 | 167.10 |
| MM-dynamic (proposed) | ✓ | ✓ | 0.3623 | 0.6377 | 112.89 [∗] |

To perform multimodal learning, the extracted features from the two networks were considered together via concatenation. The concatenated vector for multimodal learning or the output of each network for unimodal learning was input into a two-layer long short-term memory (LSTM) unit with 256 cells per layer. The weights of the LSTM layers were initialized via a normalized initialization process [21]. At the last part of the network architecture before the fully-connected layer, we applied dropout with a probability of 0.5.

*B. Dynamic Weights of Joint Loss*

In order to compare the prediction performance of the proposed model with the golden standards set by taking the majority of the annotators' evaluations, a joint loss approach was used. Joint loss is computed as the summation of the negative log-likelihood of each task.

The objective of the multimodal network is to minimize joint loss ($L_{joint}$), which combines the losses of emotion ($L_e$) and gender ($L_g$) predictions. As stated in Section 2, the way in which these losses are combined as weighted averages is important, and the key is to choose the proper weights for each loss. In [22], the authors proposed a simple method for calculating the dynamic weights of joint loss in a multi-task manner. They used this method to minimize the loss of each individual task (three different systems). The model they proposed provided different combinations of parameters for each task. In our research, we applied this concept to minimize joint loss instead of individual losses. We introduced two balancing factors, $\lambda_e$ and $\lambda_g$, which can create a balance between emotion classification and gender classification, respectively. These balancing factors are updated to determine the minimum $L_{joint}$ via backpropagation. The weights of the losses for each task, $w_e$ and $w_g$, are computed using the balancing factors as shown in (1) and (2), respectively. They are the outputs of the softmax function.

$$w_e = \frac{\exp(\lambda_e)}{\exp(\lambda_e)+\exp(\lambda_g)} \quad (1)$$

$$w_g = \frac{\exp(\lambda_g)}{\exp(\lambda_e)+\exp(\lambda_g)} \quad (2)$$

Then, the joint loss can be calculated as follows:

$$L_{joint} = w_e \times L_e + w_g \times L_g, \quad (3)$$

where the values for $w_e$ and $w_g$ vary between 0 and 1 such that $w_e + w_g = 1$. In [22], $w_e$ and $w_g$ varied in steps of 0.1, but there were no step conditions in our approach.

V. EXPERIMENTS AND RESULTS

This section presents our experiments for verifying the superiority of the proposed method. Two experiments were conducted in this research. The first experiment consisted in comparing multimodal approaches using either static weights (MM-static), which is the baseline approach [11], and dynamic joint loss weights (MM-dynamic). The second experiment consisted in a comparison between unimodal approaches and multimodal approaches for emotion and gender recognition. For these unimodal and multimodal approaches, we used dynamic joint loss weights because dynamic weights showed a better performance than static ones. The unimodal approaches had two modalities: audio (UM-dynamic-A) and video (UM-dynamic-V).

To verify the validity of the proposed method, we present our experimental results for improvised dialog from IEMOCAP data. Scripted play was not considered in our experiments because it causes greater uncertainty when assessing accurate emotions than improvised dialog [1]. The original data was divided as 6:2:2 into training, validation, and test datasets, respectively. The values of the dynamic joint loss weights, $\lambda_e$ and $\lambda_g$, were initialized as 0 for each task. It follows that $w_e$ and $w_g$ were initialized as the same values as in the baseline method, 0.5.

Tab. I shows the overall performance obtained in the experiments and illustrates the benefits of the approach that uses dynamic joint loss weights, which attained a much better performance than the approach that uses static joint loss weights (in which the weights are uniformly distributed). In addition, the performance of our proposed method surpassed the performance of the baseline approach. The final weights of the MM-dynamic method after training were 0.3623 for the emotion recognition task and 0.6377 for the gender recognition task. Even though the task weights of the UM-dynamic-V and MM-dynamic methods were the same, the total loss of the MM-dynamic method was lower. This phenomenon was caused by a multimodal effect. These results can be regarded as an experimental verification of the method proposed in Section IV.

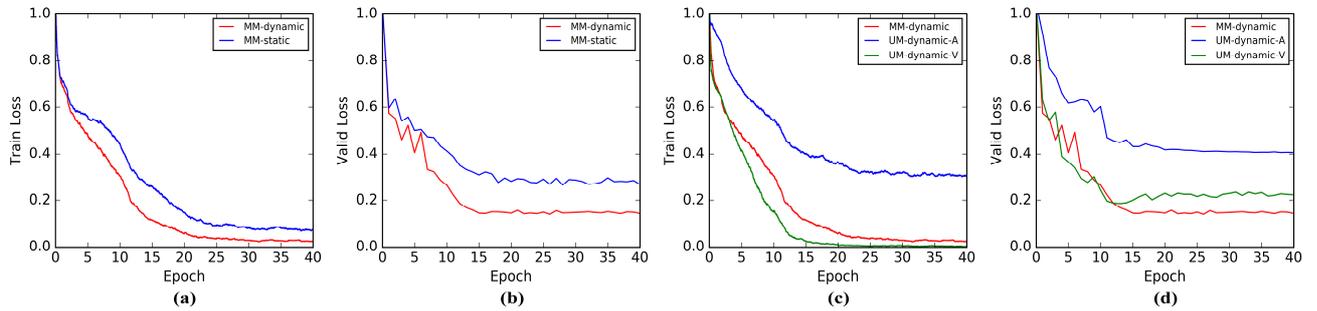

Figure 2. Graphs of joint loss versus epoch for each approach for the training and validation datasets

Fig. 2 shows the variation in joint loss from the training and validation datasets using dynamic weights and static weights for the two tasks. Figs. 2 (a) and (b) show that using dynamic weights results in better performance compared with the static approach for the training and validation datasets. Figs. 2 (c) and (d) show the results of our experiment for comparing a unimodal approach (audio and video) and a multimodal approach using dynamic joint loss weights. For the training dataset, the UM-dynamic-V method showed a slightly better performance than the MM-dynamic method. However, for the validation dataset, the loss converged at a much higher value than for the MM-dynamic method. It is clear that the UM-dynamic-V method was overfitted (showing low generalizability) and that the MM-dynamic method showed much better generalizability. Therefore, our proposed approach provided the lowest joint loss among all the approaches considered for the validation dataset.

## VI. Conclusion

In this paper, we proposed a novel end-to-end multimodal emotion and gender recognition model with dynamic joint loss weights. The proposed model does not need any pre-trained features from audio and visual data. Moreover, a dynamic approach for selecting the weights of joint loss based on the concept of multi-task learning was applied to find the proper weight for each loss. With the results of our experiments, we showed that the performance of the proposed model for emotion and gender recognition is better than that of the baseline approach using static joint loss weights. These results indicate that our proposed model has higher generalizability than other approaches. To the best of our knowledge, this research is the first to demonstrate the strength of using dynamic weights for minimizing joint loss in emotion and gender recognition tasks.

Some supplementary points should be considered to achieve better emotion and gender recognition in the future. Our proposed model should be validated using other datasets besides IEMOCAP. Additionally, the technique we used for combining multimodal data in our research was to simply concatenate the outputs of the audio and visual networks. Various approaches for multimodal learning can be applied to consider the characteristics of each modality more effectively.

## References

[1] B. W. Schuller, "Speech emotion recognition: two decades in a nutshell, benchmarks, and ongoing trends," in *Communications of the ACM*, vol. 61, no. 5, 2018, pp. 90–99.

[2] P. Ekman and W. V. Friesen, "Constants across cultures in the face and emotion," in *Journal of Personality and Social Psychology*, vol. 17, no. 2, pp. 124-129.

[3] J. A. Russell, "A circumplex model of affect," in *Journal of Personality and Social Psychology*, vol. 39, no. 6, 1980, pp. 1161–1178.

[4] R. Werber, V. Barrielle, C. Soladié, and R. Séguier, "High-level geometry-based features of video modality for emotion prediction," in *Proceedings of the 6th International Workshop on Audio/Visual Emotion Challenge*, Amsterdam, The Netherlands, October, 2016, pp. 51–58.

[5] K. Brady, Y. Gwon, P. Khorrami, E. Godoy, W. Campbell, C. Dagli, and T. S. Huang, "Multi-modal audio, video and physiological sensor learning for continuous emotion prediction," in *Proceedings of 6th International Workshop on Audio/Visual Emotion Challenge*, Amsterdam, The Netherlands, October, 2016, pp. 97–104.

[6] F. Povolny, P. Matejka, M. Hradis, A. Popková, L. Otrusina, P. Smrz, I. Wood, C. Robin, and L. Lamel, "Multimodal emotion recognition for AVEC 2016 challenge," in *Proceedings of 6th International Workshop on Audio/Visual Emotion Challenge*, Amsterdam, The Netherlands, October, 2016, pp. 75–82.

[7] A. M. Burton, V. Bruce, and N. Dench, "What's the difference between men and women? Evidence from facial measurement," in *Perception*, vol. 22, 1993, pp. 153–176.

[8] B. I. Fagot and M. D. Leinbach, "Sex-role development in young children: From discrimination to labeling," in *Developmental Review*, vol. 13, 1993, pp. 205–224.

[9] P. Tzirakis, J. Zhang, and B. W. Schuller, "End-to-end speech emotion recognition using deep neural networks," in *IEEE International Conference on Acoustics, Speech and Signal Processing (ICASSP) 2018*, Calgary, Canada, April, 2018, pp. 5089–5093.

[10] B. C. Ko, "A brief review of facial emotion recognition based on visual information," in *Sensors*, vol. 18, no. 2, 2018, 401.

[11] P. Tzirakis, G. Trigeorgis, M. A. Nicolaou, B. W. Schuller, and S. Zafeiriou, "End-to-end multimodal emotion recognition using deep neural networks," in *IEEE Journal of Selected Topics in Signal Processing*, vol. 11, no. 8, 2017, pp. 1301–1309.

[12] J. Kim, G. Englebienne, K. P. Truong, and V. Evers, "Towards speech emotion recognition in the wild using aggregated corpora and deep multi-task learning," in *Interspeech 2017*, Stockholm, Sweden, August, 2017, pp. 1113–1117

[13] D. Le, Z. Aldeneh, and E. M. Provost, "Discretized continuous speech emotion recognition with multi-task deep recurrent neural network," in *Interspeech 2017*, Stockholm, Sweden, August 2017, pp. 1108–1112.

[14] G. Trigeorgis, F. Ringeval, R. Brueckner, E. Marchi, M. A. Nicolaou, B. W. Schuller, and S. Zafeiriou, "Adieu features? End-to-end speech emotion recognition using a deep convolutional recurrent neural network," in *IEEE International Conference on Acoustics, Speech and Signal Processing (ICASSP) 2016*, Shanghai, China, 2016.

[15] Z. Huang, B. Stasak, T. Dang, K. W. Gamage, P. Le, V. Sethu, and J. Epps, "Staircase regression in OA RVM, data selection and gender dependency in AVEC 2016," in *Proceedings of 6th International Workshop on Audio/Visual Emotion Challenge*, Amsterdam, The Netherlands, October, 2016, pp. 19–26.

[16] J. Han, Z. Zhang, N. Cummins, F. Ringeval, and B. W. Schuller, "Strength modelling for real-world automatic continuous affect recognition from audiovisual signals," in *Image and Vision Computing*, vol. 65, 2017, pp. 76–86.

[17] N. K. Kim, J. Lee, H. K. Ha, G. W. Lee, J. H. Lee, and H. K. Kim, "Speech emotion recognition based on multi-task learning using a convolutional neural network," in *IEEE Asia-Pacific Signal and Information Processing Association Annual Summit and Conference (APSIPA ASC) 2017*, Kuala Lumpur, Malaysia, December, 2017, pp. 704–707.

[18] C. Busso, M. Bulut, C. C. Lee, A. Kazemzadeh, E. Mower, S. Kim, J. N. Chang, S. Lee, and S. S. Narayanan, "IEMOCAP: Interactive emotional dyadic motion capture database," in *Journal of Language Resources and Evaluation*, vol. 42, no. 4, 2008, pp. 67–80.

[19] M. Abdelwahab, and C. Busso, "Domain adversarial for acoustic emotion recognition," in *IEEE Transactions on Audio, Speech, and Language Processing*, submitted for publication.

[20] K. He, X. Zhang, S. Ren, and J. Sun, "Deep residual learning for image recognition," in *Proceedings of the IEEE Conference on Computer Vision and Pattern Recognition (CVPR)*, Las Vegas, USA, July, 2016, pp. 770–778.

[21] X. Glorot and Y. Bengio, "Understanding the difficulty of training deep feedforward neural networks," in *Proceedings of the 13th International Conference on Artificial Intelligence and Statistics (AISTATS)*, Sardinia, Italy, March, 2010, pp. 249–256.

[22] S. Parthasarathy and C. Busso, "Jointly predicting arousal, valence, and dominance with multi-task learning," in *Interspeech 2017*, Stockholm, Sweden, August, 2017, pp. 1103–1107.